# Evaluation eines neuen Ansatzes zur automatischen Volumenbestimmung von Glioblastomen anhand von mehreren manuellen Expertensegmentierungen

# Evaluation of a Novel Approach for Automatic Volume Determination of Glioblastomas Based on Several Manual Expert Segmentations

J. Egger [1, 2], M. H. A. Bauer [1, 2], D. Kuhnt [1], B. Carl [1], C. Kappus [1], B. Freisleben [2], C. Nimsky [1]
[1]Philipps-Universität Marburg, Klinik für Neurochirurgie, Marburg, Deutschland
[2]Philipps-Universität Marburg, Fachbereich Mathematik und Informatik, Marburg, Deutschland

egger@med.uni-marburg.de

## Kurzfassung

Das Glioblastom ist der häufigste maligne, hirneigene Tumor und zählt zu den höchst-malignen Neoplasien des menschlichen Organismus. Zum klinischen Follow-up ist die Evaluation des Tumorvolumens während des weiteren Krankheitsverlaufs essentiell. Die manuelle Segmentierung zur Erhebung des Tumorvolumens ist jedoch ein zeitaufwändiger Prozess. In diesen Beitrag wird ein neuer Ansatz zur automatischen Segmentierung und Volumenbestimmung von Glioblastomen (Glioblastoma multiforme) vorgestellt und evaluiert. Der Ansatz verwendet einen benutzerdefinierten Saatpunkt innerhalb des Glioms, um einen gerichteten 3D Graphen aufzubauen. Die Knoten des Graphen werden dabei durch Abtasten von Strahlen gewonnen, die durch die Oberflächenpunkte eines Polyeders verlaufen. Nachdem der Graph konstruiert wurde, wird der minimale s-t-Schnitt berechnet, der das Glioblastom vom Hintergrund trennt. Zur Evaluation wurden 12 Magnetresonanztomographie (MRT) Daten von Neurochirurgen mit mehreren Jahren Erfahrung in der Resektion von Gliomen manuell Schicht-für-Schicht segmentiert. Anschließend wurden diese manuellen Segmentierungen anhand des Dice Similarity Koeffizienten (DSC) mit den Ergebnissen des vorgestellten Ansatzes verglichen. Um das Ergebnis des DSC besser einschätzen zu können, wurden die manuellen Segmentierungen der Experten auch untereinander verglichen und anhand des DSC ausgewertet. Zusätzlich wurden die 12 Datensätze von einem Neurochirurgen nach einem Zeitraum von 2 Wochen ein weiteres Mal segmentiert, um auch hier die Abweichung des DSC zu messen.

## Abstract

The glioblastoma multiforme is the most common malignant primary brain tumor and is one of the highest malignant human neoplasms. During the course of disease, the evaluation of tumor volume is an essential part of the clinical follow-up. However, manual segmentation for acquisition of tumor volume is a time-consuming process. In this paper, a new approach for the automatic segmentation and volume determination of glioblastomas (glioblastoma multiforme) is presented and evaluated. The approach uses a user-defined seed point inside the glioma to set up a directed 3D graph. The nodes of the graph are obtained by sampling along rays that are sent through the surface points of a polyhedron. After the graph has been constructed, the minimal s-t cut is calculated to separate the glioblastoma from the background. For evaluation, 12 Magnetic Resonance Imaging (MRI) data sets were manually segmented slice by slice, by neurosurgeons with several years of experience in the resection of gliomas. Afterwards, the manual segmentations were compared with the results of the presented approach via the Dice Similarity Coefficient (DSC). For a better assessment of the DSC results, the manual segmentations of the experts were also compared with each other and evaluated via the DSC. In addition, the 12 data sets were segmented once again by one of the neurosurgeons after a period of two weeks, to also measure the intra-physician deviation of the DSC.

## 1 Einleitung

Als Gliome werden Tumore bezeichnet, die von den Stützzellen des Gehirns ausgehen. Sie sind die häufigsten hirneigenen Tumoren. Die Entität richtet sich nach der Ursprungszelle. Somit sind Astrozytome ausgehend von Astrozyten, Oligodendrogliome von Oligodendrozyten und Ependymome von Ependymzellen. Zudem existieren Mischformen dieser histopathologischen Subtypen, wie z.B. Oligoastrozytome. Astrozytome sind mit über 60% die häufigsten Gliome. Die Klassifikation erfolgt nach der World Health Organisation (WHO). Hier werden vier Subtypen unterschieden (I-IV), wobei Grad I Tumore als wenig aggressiv und proliferativ einzustufen sind [1]. Über 70% zählen zu den malignen Gliomen – WHO Grad III (anaplastisches Astrozytom) und IV (Glioblastoma multiforme, Bild 1). Den Eigennamen Glioblastoma multiforme (GBM) erhielt der Grad IV Tumor aufgrund seiner histopathologischen Erscheinung.

Das Glioblastom ist der häufigste maligne hirneigene Tumor und zählt zu den höchst-malignen Neoplasien des menschlichen Organismus. Das interdisziplinäre Behandlungskonzept vereint heute die maximale mikrochirurgische Resektion, gefolgt von perkutaner Radiatio und zumeist Chemotherapie. Neue Bestrahlungskonzepte sowie die Etablierung der Alkylanzien (z.B. Temozolomid) als potentes Chemotherapeutikum konnte die Überlebensrate bis dato auf lediglich ca. 15 Monate steigern [2].

Bis vor wenigen Jahren war die Rolle der chirurgischen Tumorvolumenreduktion umstritten, wobei nun bereits von mehreren Autoren ein Zusammenhang zwischen der prozentual resezierten Tumormasse und der Überlebenszeit des Patienten belegt werden konnte [3]. Das operative Prozedere wird heute durch die sogenannte Neuronavigation optimiert. Zudem findet die Integration funktioneller Daten wie Diffusions-Tensor-Bildgebung (DTI), funktionellem MRT (fMRI), Magnetencephalographie (MEG) oder Magentresonanzspektroskopie (MRS) und Positronemissionstomographie (PET) Anwendung. Zum klinischen Follow-up ist die Evaluation des Tumorvolumens während des weiteren Krankheitsverlaufs essentiell. Die manuelle Segmentierung zur Erhebung des Tumorvolumens ist jedoch ein zeitaufwändiger Prozess.

In diesem Beitrag wird ein neuer Ansatz zur Segmentierung von WHO Grad IV Gliomen vorgestellt. Hierbei wird einen benutzerdefinierter Saatpunkt innerhalb des Tumors genutzt, um einen gerichteten 3D-Graphen aufzubauen. Auf dem Graphen wird dann ein minimaler s-t-Schnitt berechnet, der das Glioblastom vom Hintergrund trennt. Evaluiert wurde der Ansatz anhand von 12 MRT Aufnahmen, die von Experten manuell segmentiert wurden.

Der Beitrag ist wie folgt aufgebaut: Abschnitt 2 stellt den Stand der Forschung dar. Abschnitt 3 präsentiert den neuen Ansatz. In Abschnitt 4 werden experimentelle Ergebnisse diskutiert. Abschnitt 5 fasst den Beitrag zusammen und gibt einen Ausblick auf zukünftige Arbeiten.

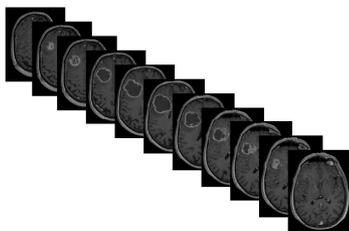

**Bild 1** Mehrere axiale Schichten einer kontrastierten T1-gewichteten MRT Aufnahme eines Patienten mit einem Glioblastom (WHO Grad IV).

## 2 Stand der Forschung

In der Literatur finden sich mehrere Algorithmen für die (semi-)automatische Segmentierung von Gliomen aus MRT Aufnahmen. Eine gute Übersicht der deterministischen und statistischen Ansätze liefert die Publikation von Angelini et al. [4]. Die meisten Ansätze sind regionenbasiert, aktuelle Ansätze basieren auf deformierbaren Modellen, die Kanteninformationen mit einbeziehen.

Gibbs et al. [5] stellen eine Kombination aus einem Regionenwachstumsverfahren und morphologischer Kantendetektion für die Segmentierung von Tumoren in T1 gewichteten MRT Aufnahmen vor. Basierend auf einer manuellen Vorgabe des Tumorsignals und der umgebenden Struktur wird eine Segmentierung ausgeführt, die Pixel-Schwellwerte, morphologisches Opening und Closing und das Anpassen an eine Kanten-Map nutzt. Die Autoren haben ihren Ansatz mit einem Phantomdatensatz und zehn klinischen Datensätzen evaluiert. Allerdings lag die durchschnittliche Zeit für eine Segmentierung bei zehn Minuten und die Gehirntumore, die sie segmentierten, waren nicht genau definiert. Eine interaktive Methode zur Segmentierung von *full-enhancing*, *ring-enhancing* und *non-enhancing* Tumoren wurde von Letteboer et al. [6] vorgestellt. Die Methode wurde anhand von zwölf Datensätzen evaluiert. Basierend auf einem manuellen Tracing einer initialen Schicht wurden mehrere morphologische Filteroperationen auf dem MRI Volumen angewendet, um die Daten in homogene Bereiche zu separieren. Aufbauend auf intensitäts-basierten Wahrscheinlichkeiten für Tumorgewebe präsentierten Droske et al. [7] ein deformierbares Modell, das eine Level Set Formulierung benutzt, um die MRT-Daten in Regionen mit ähnlichen Bildeigenschaften für die Tumorsegmentierung zu unterteilen. Diese modelbasierte Segmentierung wurde anhand von zwölf Patientendaten getestet. Clark et al. [8] stellen eine wissensbasierte, automatische Segmentierung von multispektralen Daten vor, um Glioblastome einzuteilen. Nach einer Trainingsphase mit Fuzzy C-means Klassifikation, einer Cluster Analyse und der Berechnung einer Gehirnmaske, wird eine initiale Tumorsegmentierung anhand von Histogrammschwellwerten ausgeführt, um Nicht-Tumorpixel zu eliminieren. Das präsentierte System wurde mit drei Volumendatensätzen trainiert und dann auf dreizehn neuen Datensätzen getestet. Eine Segmentierung basierend auf Outlier-Erkennung in T2 gewichteten MR Daten wurde von Prastawa et al. [9] durchgeführt. Die Bilddaten werden hierbei auf einen normalen Gehirnatlas registriert, um abnormale Tumorregionen zu erkennen. Der Tumor und das Ödem werden dann anhand von statistischem Clustering der verschiedenen Voxel und einem deformierbaren Modell isoliert. Allerdings haben die Autoren den Ansatz nur auf drei Datensätzen getestet, und für jede Segmentierung benötigte der Algorithmus ungefähr 90 Minuten. Sieg et al. [10] stellten eine Methode vor, um kontrastmittelverstärkte Gehirntumore und anatomische Strukturen aus registrierten, multispektralen Daten zu segmentieren. Dazu wurden neuronale multilayer-feedforward Netzwerke mit Backpropagation trainiert und eine pixelorientierte Klassifikation für die Segmentierung angewendet. Der Ansatz wurde anhand von 22 Datensätzen getestet, allerdings wurden keine Rechenzeiten angegeben.

## 3 Methoden

Das vorgestellte Verfahren lässt sich in zwei Schritte unterteilen: In einem ersten Schritt wird von einem benutzerdefinierten Saatpunkt aus ein gerichteter 3D Graph aufgebaut. In einem zweiten Schritt wird der minimale s-t

Schnitt auf diesem Graphen berechnet und somit der Tumor vom Hintergrund getrennt. Die einzelnen Schritte werden in den Abschnitten 3.1 und 3.2 vorgestellt.

### 3.1 Aufbau des Graphen

Der Graph wird ausgehend von einem benutzerdefinierten Saatpunkt aufgebaut, der innerhalb des Tumors liegt. Die Knoten des Graphen werden durch Abtasten von Strahlen gewonnen, die durch die Oberflächenpunkte eines Polyeders verlaufen (Mittelpunkt des Polyeders ist Saatpunkt). Die abgetasteten Punkte sind die Knoten $n \in V$ vom Graphen $G(V,E)$ und $e \in E$ ist eine Satz von Kanten. Es gibt zwei Arten von Kantentypen: Kanten, die den Graphen mit einer Quelle $s$ und einer Senke $t$ verbinden und Kanten innerhalb des Graphen. Bei den Kanten innerhalb des Graphen gibt es wiederum mehrere Arten.

Die Kanten $<v_i, v_j> \in E$ des Graphen $G$ verbinden immer zwei Knoten $v_i, v_j$ innerhalb des Graphen. Dabei gibt es unter anderen zwei ∞-gewichtete Kanten: z-Kanten $A_z$ und r-Kanten $A_r$ ($Z$ ist die Anzahl der abgetasteten Punkte entlang eines Strahls $z=(0,...,Z-1)$ und $R$ ist die Anzahl der Strahlen, die durch die Oberflächenpunkte des Polyeders gesendet werden $r=(0,...,R-1)$, wobei $V(x_n,y_n,z_n)$ ein Nachbarpunkt von $V(x,y,z)$ ist):

$$A_z = \{\langle V(x,y,z), V(x,y,z-1)\rangle \mid z > 0\}$$
$$A_r = \{\langle V(x,y,z), V(x_n,y_n,\max(0,z-\Delta_r))\rangle\}$$

Die Kanten zwischen zwei Knoten entlang eines Strahls $A_z$ stellen sicher, dass alle Knoten unterhalb der Polyederoberfläche im Graphen in einem *closed set* enthalten sind. Die Kanten $A_r$ zwischen den Knoten der unterschiedlichen Strahlen schränken die Anzahl der möglichen Segmentierungen ein und erzwingen eine Glattheit der resultierenden Oberfläche mit Hilfe eines Parameters $\Delta_r$. Je größer $\Delta_r$ ist, desto mehr mögliche Segmentierungen gibt es (Bild 2).

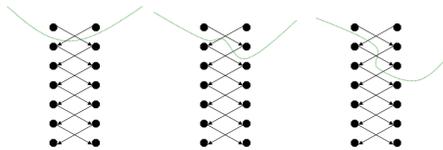

**Bild 2** Prinzip der Kantenschnitte bei zwei Strahlen für $\Delta_r=1$. Gleiche Kosten (2·∞) für einen Schnitt (linkes und mittleres Bild). Höhere Kosten (4·∞) für einen Schnitt (rechtes Bild).

### 3.2 Minimaler s-t Schnitt

Nach der Graphkonstruktion wird das *closed set* des Graphen mit minimalen Kosten anhand eines s-t-Schnittes berechnet [11]. Dieser liefert eine optimale Segmentierung des Tumors unter dem Einfluss des Parameters $\Delta_r$, der die Steifigkeit der Oberfläche beeinflusst. Ein Deltawert von 0 stellt sicher, dass das Segmentierungsergebnis eine Kugel ist. Die Kosten $w(x, y, z)$ für die Kanten $v \in V$ zur Quelle und Senke werden folgendermaßen berechnet: Gewichte haben einen Wert von $c(x,y,z)$, wenn $z$ Null oder maximal ist, ansonsten $c(x,y,z)-c(x,y,z-1)$, wobei $c(x,y,z)$ der Betrag der Differenz zwischen einem durchschnittlichen Grauwert des Tumors und dem Grauwert des Voxels an Position $(x,y,z)$ ist. Der durchschnittliche Grauwert zur Berechnung der Kosten ist essentiell für das Segmentierungsergebnis. Basierend auf der Annahme, dass der benutzerdefinierte Saatpunkt innerhalb des Tumors sitzt, kann der durchschnittliche Grauwert allerdings automatisch bestimmt werden. Dazu wird über eine Region der Dimension $d$ um den benutzerdefinierten Saatpunkt $(s_x, s_y, s_z)$ integriert:

$$\int_{-d/2}^{d/2}\int_{-d/2}^{d/2}\int_{-d/2}^{d/2} T(s_x+x, s_y+y, s_z+z)\,dx\,dy\,dz$$

## 4 Ergebnisse

Die Methoden wurden in C++ innerhalb der medizinischen Plattform MeVisLab realisiert (http://www.mevislab.de). Eine komplett automatische Segmentierung – beginnend mit der Graphkonstruktion bis hin zur Berechnung des Min Cut – benötigte in unserer Implementierung weniger als 5 Sekunden (gemessen auf einem Intel Core i5-750 CPU, 4x2.66 GHz, 8 GB RAM, Windows XP Professional x64 Version, 2003, SP 2). Eine manuelle Segmentierung dagegen dauerte bei den Experten 8.03±5.18 min. (A), 5.22±2.66 min (B) bzw. 5.4±2.07 min. (C) (Mittelwert $\mu$ ± Standardabweichung $\sigma$).

Für die Evaluierung wurden 12 T1-gewichtete MRT Aufnahmen mit WHO Grad IV Gliomen aus der klinischen Routine verwendet. Die manuellen Segmentierungen wurden sowohl untereinander, als auch mit den automatischen Segmentierungen anhand des Dice Similarity Koeffizienten (DSC) ausgewertet [12].

In Tabelle 1 ist der Vergleich der automatischen Segmentierungsergebnisse mit den manuellen Segmentierungen von drei Neurochirurgen (A, B, C) zu sehen. Neben dem minimalen und maximalen DSC sind Mittelwert $\mu$ und Standardabweichung $\sigma$ angegeben. Um den Vergleich der automatischen und manuellen Ergebnisse aus Tabelle 1 besser einschätzen zu können, wurden die Ergebnisse von B und C mit A verglichen, sowie die Glioblastome von A zweimal segmentiert (Tabelle 2).

In Bild 3 ist das Ergebnis einer automatischen Segmentierung eines Glioblastoms dargestellt. Auf der linken Seite ist die Volumenmaske in den MRT Datensatz eingeblendet. Bild 4 zeigt zwei Segmentierungsergebnisse und die Abweichung eines Neurochirurgen für eine axiale Schicht. Die Segmentierungen wurden von einem Neurochirurgen innerhalb von zwei Wochen durchgeführt.

|  | manuelle Segmentierungen (DSC) | | |
|---|---|---|---|
|  | A | B | C |
| min | 69.82% | 65.62% | 58,66% |
| max | 93.82% | 91.78% | 90,96% |
| $\mu \pm \sigma$ | 79.96±8.06% | 77.79±8.49% | 76,83±13,67% |

**Tabelle 1** Vergleich der automatischen Segmentierungsergebnisse mit den manuellen Segmentierungsergebnissen von drei Neurochirurgen (A, B, C).

|   | manuelle Segmentierungen (DSC) | | |
|---|---|---|---|
|   | A | B | C |
| min | 84.01% | 78.68% | 76,03% |
| max | 96.30% | 94.86% | 94,83% |
| $\mu \pm \sigma$ | 90.29 ± 4.48% | 88 ± 6.08% | 86,63 ± 6,87% |

**Tabelle 2** Vergleich der manuellen Segmentierung des Neurochirurgen A mit den manuellen Segmentierungsergebnissen von Neurochirurgen A, B und C.

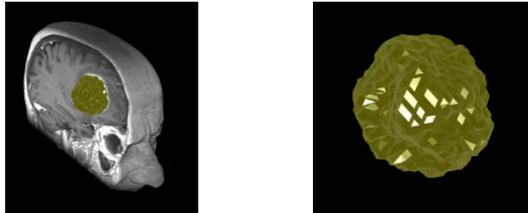

**Bild 3** Automatisch segmentiertes Glioblastom (WHO Grad IV) eingeblendet in eine kontrastierte T1-gewichteten MRT Aufnahme eines Patienten (links). Volumenmaske des segmentierten Glioblastoms (rechts).

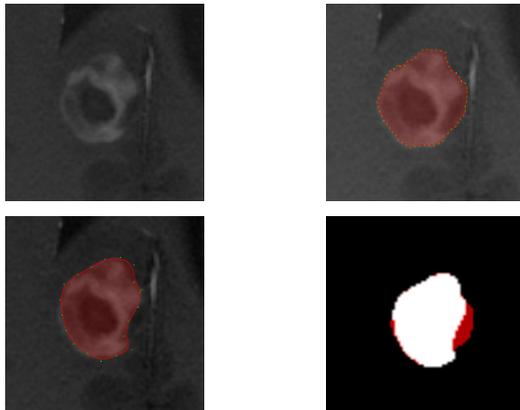

**Bild 4** Axiale Schicht einer kontrastierten T1-gewichteten MRT Aufnahme eines Patienten mit einem Glioblastom (links oben). Manuelles Segmentierungsergebnis eines Neurochirurgen (rechts oben). Manuelles Segmentierungsergebnis desselben Neurochirurgen zwei Wochen später (links unten). Übereinandergelegte Segmentierungsergebnisse (rechts unten).

## 5 Zusammenfassung und Ausblick

In diesem Beitrag wurde ein neuer Ansatz zur Segmentierung von WHO Grad IV Gliomen vorgestellt. Der Ansatz nutzt einen benutzerdefinierten Saatpunkt innerhalb des Tumors, um einen gerichteten 3D-Graphen aufzubauen. Auf dem Graphen wird dann ein minimaler s-t-Schnitt berechnet, der das Glioblastom vom Hintergrund trennt. Evaluiert wurde der Ansatz anhand von 12 MRT Aufnahmen, die von Experten mit mehreren Jahren Erfahrung in der Entfernung von Gehirntumoren manuell Schicht-für-Schicht segmentiert wurden. Um die Abweichungen zwischen den Segmentierungsergebnissen und den manuellen Segmentierungen besser einschätzen zu können, wurden die manuellen Segmentierungen der Neurochirurgen auch untereinander verglichen. Außerdem wurden die 12 Gliome von einem Arzt mehrmals innerhalb von zwei Wochen segmentiert, um auch hier die Abweichung zu messen.

In einem nächsten Schritt soll das vorgestellte Verfahren auf zerebrale Aneurysmen und Hypophysenadenome angewendet werden. Dabei kommt es vor allem auf den Verlauf nach einem Eingriff an. Das Volumen eines zerebralen Aneurysmas oder eines Hypophysenadenoms soll in jeder Follow-up Aufnahme automatisch berechnet und gespeichert werden. Dadurch ist es möglich – zwischen mehreren Follow-up Scans – auch sehr geringe Größenänderungen automatisch und sehr genau zu detektieren und dem behandelnden Arzt anzuzeigen.

## 6 Literatur